\documentclass[11pt]{article}
\usepackage{acl}

\usepackage{booktabs}
\usepackage{multirow}
\usepackage{makecell}
\PassOptionsToPackage{table}{xcolor}
\usepackage{xcolor}
\usepackage{colortbl}
\usepackage{listings}
\usepackage{stfloats}
\usepackage{float}
\usepackage{array}
\usepackage{amsmath}
\usepackage{arydshln}

\usepackage{pifont}
\newcommand{\cmark}{\ding{51}}
\newcommand{\xmark}{\ding{55}}

\definecolor{codebg}{rgb}{0.96, 0.96, 0.96}
\definecolor{tagblue}{rgb}{0.0, 0.3, 0.7}
\definecolor{citered}{rgb}{0.8, 0.1, 0.1}
\definecolor{stringgreen}{rgb}{0.1, 0.5, 0.1}
\definecolor{turngray}{rgb}{0.4, 0.4, 0.4}
\definecolor{deepblue}{HTML}{1565C0}

\usepackage{times}
\usepackage{latexsym}
\usepackage[T1]{fontenc}
\usepackage[utf8]{inputenc}
\usepackage{microtype}
\usepackage{inconsolata}
\usepackage{graphicx}
\usepackage{fontawesome5}

\title{ReCite: Agentic Reasoning for Faithful Citation}

\author{
  \textbf{Yuyang Huang\textsuperscript{1}},
  \textbf{Bobo Li\textsuperscript{2}\thanks{Corresponding authors.}},
  \textbf{Jiajia Song\textsuperscript{2}},
  \textbf{Yuzhe Ding\textsuperscript{1}},
  \textbf{Chong Teng\textsuperscript{1}},
  \textbf{Fei Li\textsuperscript{1}},
  \textbf{Donghong Ji\textsuperscript{1}\footnotemark[1]}
\\
  \textsuperscript{1}School of Cyber Science and Engineering, Wuhan University\enspace
  \textsuperscript{2}National University of Singapore \\
  \texttt{\{hyy279, dhji\}@whu.edu.cn},\enspace
  \texttt{\{libobo, jjiasong\}@nus.edu.sg} \\
  \textcolor{deepblue}{\faGlobe\enspace\url{https://hyy279.github.io/ReCite/}}
}

\begin{document}
\maketitle
\begin{abstract}
Accurate citations are the foundation of academic writing, tracing intellectual origins and substantiating core claims. However, manually navigating the growing volume of scientific literature is increasingly difficult, prompting reliance on automatic citation recommendation. While modern retrieval-augmented architectures have largely mitigated the fabrication of non-existent papers, current systems relying on semantic similarity struggle with misattribution, often citing authentic papers that fail to logically support the author's claim. To address this challenge, we argue that accurate citation requires a shift from similarity-based search to active, claim-level reasoning. We propose ReCite, a decoupled agentic framework that orchestrates location perception, intent-aware query planning, and reflective verification.
Trained on synthesized reasoning trajectories, our agent verifies claim-evidence consistency and triggers self-correction loops when retrieved candidates lack logical support.
Experiments demonstrate that our lightweight framework outperforms state-of-the-art massive generative models in strict citation accuracy. By grounding literature matching in verifiable logic rather than semantic overlap, ReCite establishes a reliable foundation for automated academic writing.
\end{abstract}

\section{Introduction}

Accurate citations are the foundation of academic writing, serving to trace intellectual origins and substantiate core claims.
However, as the volume of scientific literature grows exponentially, manually tracking relevant prior work and locating the most appropriate papers have become increasingly difficult~\cite{bornmann2015growth, fortunato2018science}.
To alleviate this growing cognitive burden, automatic citation recommendation (ACR) has been widely adopted as an essential research tool~\cite{gao-etal-2023-enabling}.
As large language models (LLMs) increasingly draft academic texts, they introduce a severe risk of generating plausible but incorrect references~\cite{ji23survey, sakai26arr}, forcing ACR to evolve from a passive writing aid into an active quality gatekeeper.

\begin{figure}[!t]
    \centering
    \includegraphics[width=\linewidth]{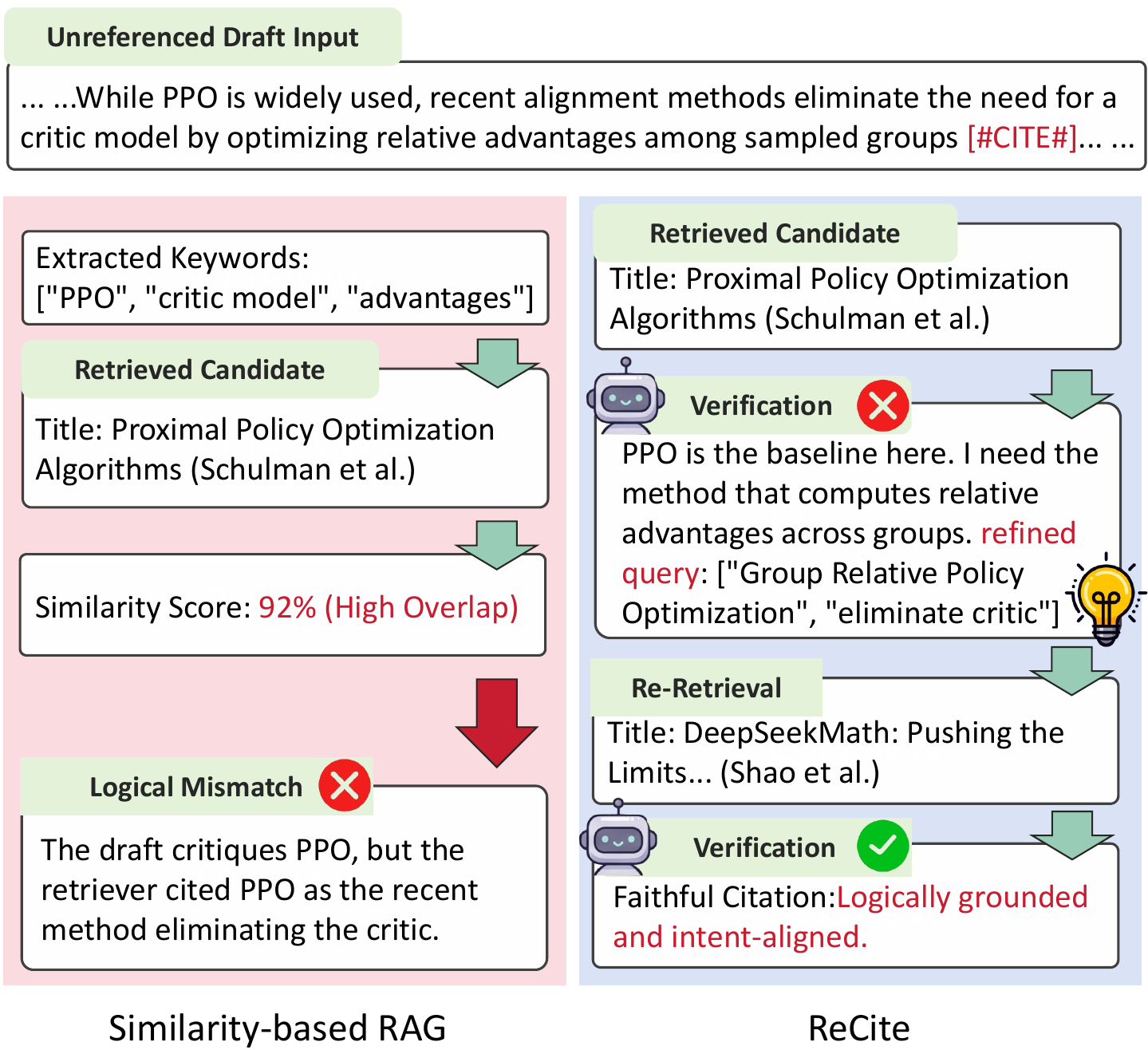}
    \caption{Comparison of citation paradigms: Traditional semantic retrieval versus our framework.}
    \label{fig:intro}
\end{figure}

For years, citation recommendation has been modeled as a similarity search problem~\cite{gu2022local, gao-etal-2023-enabling, wang2025scholar}.
Both early retrievers~\cite{Huang2014RefSeerAC} and recent LLM-based agents~\cite{celik-tekir-2025-citebart, Zhang2026LIMRIM} equate citing with finding a textually relevant paper.
In the LLM era, citation hallucinations~\cite{algaba-etal-2025-large, sakai26arr} have crystallized into two distinct forms.
Fabrication, inventing non-existent papers, is largely resolved by retrieval-augmented architectures; misattribution, citing authentic papers that fail to logically support the claim, exposes the core limit of similarity-based retrieval and remains open~\cite{xu-etal-2025-aliice}.

To address this, we argue that citation is fundamentally a reasoning task, not a retrieval one: topical relevance does not guarantee logical support for a claim.
For example, a claim about the Transformer's success in NLP cannot be supported by the vision transformer paper~\cite{alex21vit} despite their textual overlap; ``Attention is All You Need''~\cite{ashish17nips} is the correct logical pillar.
Yet semantic retrievers~\cite{cohan-etal-2020-specter} ignore that different citation contexts, such as introducing a background versus comparing a method, demand distinct verification criteria~\cite{cohan19ssc}.
Worse, current systems operate as unidirectional pipelines~\cite{medic-snajder-2020-improved}, so a single faulty retrieval becomes an irrevocable error without opportunity for self-correction.
Together, solving misattribution requires intent-aware retrieval, claim-level verification, and a reflective loop for error recovery.

To fulfill these requirements, we propose ReCite (Figure~\ref{fig:intro}), an agentic reasoning framework for faithful citation, trained via supervised fine-tuning and reinforcement learning.
It reshapes the citation process from a linear pipeline into an iterative architecture of reasoning, retrieval, verification, and reflection.
Operating in an observe-think-act format~\cite{yao2023reactsynergizingreasoningacting}, the agent is built on Qwen3-4B~\cite{yang2025qwen3technicalreport}, with key modules reinforced via Group Relative Policy Optimization (GRPO)~\cite{shao2024deepseekmathpushinglimitsmathematical}.
Within this framework, we instantiate three core mechanisms.
First, claim-evidence verification tackles misattribution by cross-checking each candidate's metadata against the inferred citation intent, accepting a citation only upon consistent evidence.
Second, an intent-aware query planner addresses context misalignment by generating retrieval keywords conditioned on predicted citation intents.
Third, a reflective re-retrieval loop overcomes single points of failure. Upon verification failure, the agent analyzes the cause, rewrites the query, and retries.
Since retrieval queries only target authentic databases, hallucinating entirely non-existent papers is eliminated; ReCite then focuses entirely on resolving misattribution.

Experiments on the benchmark show that ReCite substantially outperforms strong generative baselines in strict citation accuracy.
Ablations confirm that claim-evidence verification, intent-aware planning, and the reflective loop are indispensable to this gain.
In summary, our main contributions are:
\begin{itemize}
    \item
We systematically decouple citation hallucinations into fabrication and misattribution, establishing misattribution as the core challenge in the LLM era and reframing citation as a reasoning task.

    \item
We propose ReCite, the first closed-loop claim-evidence verification framework, integrating intent-aware query planning and a reflective re-retrieval loop.

    \item Experiment results highlight the inherent challenge of citation reasoning for current LLMs and demonstrate the effectiveness of ReCite.
\end{itemize}

\section{Related Work}

\textbf{Local and Global Citation Recommendation.}
ACR has transitioned from global representations and two-stage pipelines \cite{beltagy-etal-2019-scibert, medic-snajder-2020-improved, gu2022local} to localized, context-aware generation \cite{celik-tekir-2025-citebart, BUSCALDI2024103583} and dynamic Retrieval-Augmented Generation (RAG) frameworks \cite{wang2025scholar, Zhang2026LIMRIM}.
Although these generative models excel at contextual integration, relying on semantic similarity as a generative prior often bypasses strict logical entailment.
Consequently, these systems frequently suffer from citation fidelity issues, including hallucinated references and misattributed claims \cite{algaba-etal-2025-large, sakai26arr, celik-tekir-2025-citebart}.

\textbf{Logical Verification and Agentic Reasoning.}
Standardized benchmarks such as ALCE \cite{gao-etal-2023-enabling} have begun to quantify citation reliability, prompting verification approaches based on Natural Language Inference (NLI) \cite{xu-etal-2025-aliice} or theorem provers \cite{Pei2025FoVerFL}.
However, such static post-hoc filtering cannot correct errors that have propagated through a unidirectional generation pipeline.
Autonomous frameworks (e.g., ReAct \cite{yao2023reactsynergizingreasoningacting}, Reflexion \cite{NEURIPS2023_1b44b878}) and step-level reinforcement (e.g., GRPO \cite{Li2026StepGRPOER}) introduce self-correction~\cite{li26taming}, but remain constrained by pre-trained priors and incur high context overhead.
Current citation datasets only map local contexts to target papers; they omit the explicit reasoning traces required to train agents for this task.
To build reasoning-driven citation models, the field requires a specialized dataset that supports location perception, intent reasoning, and closed-loop reflective trajectories.

\begin{figure}[!t]
    \centering
    \includegraphics[width=\linewidth]{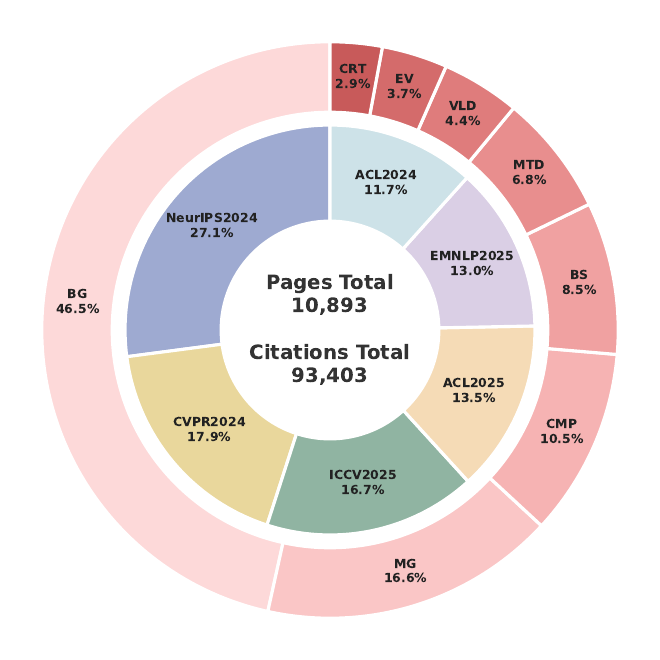}
    \caption{The proportion distribution of academic conference sources and the various citation intent categories within the utilized citation dataset.
    }
    \label{fig:data_source}
\end{figure}

\section{Dataset Construction}
\label{sec:data}

We construct a large-scale literature dataset for training an agentic citation workflow.
Given a draft paragraph, the agent must (1) detect which positions require a citation, (2) infer each citation's intent and generate targeted retrieval queries, and (3) verify whether retrieved candidates support the local claim, retrying with refined queries on failure.
The dataset therefore provides three task-specific subsets that supply the supervision for the three corresponding modules of ReCite, namely CiteLocator, QueryPlanner, and Master Brain.

\subsection{Data Source and Preparation}
\label{sec:data_prep}

Our raw corpus consists of 10,893 LaTeX source packages crawled from arXiv, corresponding to papers published in 2024--2025 at top computer science venues, including ACL, EMNLP, CVPR, ICCV, and NeurIPS.
Working at the \LaTeX{} source level preserves the explicit \texttt{\textbackslash cite} commands and their bibliography keys, giving us ground-truth citation positions and references for downstream supervision.

We apply a four-step preparation pipeline: extracting introduction sections, normalizing heterogeneous citation macros (\texttt{\textbackslash cite}, \texttt{\textbackslash citep}, \texttt{\textbackslash citet}, etc.) into a unified form, filtering non-textual noise, and removing paragraphs with abnormal citation density.
Figure~\ref{fig:data_source} shows the venue-level distribution of the resulting corpus.
Compared with existing citation datasets such as RefSeer~\cite{Huang2014RefSeerAC} and S2ORC~\cite{Lo2020S2ORCTS}, which provide only static context-to-reference mappings, our dataset additionally captures the location, intent, and reflective trajectories required to train an agentic citation workflow (Table~\ref{tab:dataset_comparison}).

\begin{table}[!t]
    \centering
    \small
    \begin{tabular}{lccc}
        \toprule
        \textbf{Features} & \textbf{RefSeer} & \textbf{S2ORC} & \textbf{Ours} \\
        \midrule
        Local Context Mapping    & \textcolor{stringgreen}{\cmark} & \textcolor{stringgreen}{\cmark} & \textcolor{stringgreen}{\cmark} \\
        Full-Text Structure      & \textcolor{citered}{\xmark} & \textcolor{stringgreen}{\cmark} & \textcolor{stringgreen}{\cmark} \\
        \hdashline \noalign{\vskip 0.5ex}
        Location Perception      & \textcolor{citered}{\xmark} & \textcolor{citered}{\xmark} & \textcolor{stringgreen}{\cmark} \\
        Intent Reasoning   & \textcolor{citered}{\xmark} & \textcolor{citered}{\xmark} & \textcolor{stringgreen}{\cmark} \\
        Reflective Trajectories  & \textcolor{citered}{\xmark} & \textcolor{citered}{\xmark} & \textcolor{stringgreen}{\cmark} \\
        \bottomrule
    \end{tabular}
    \caption{Comparison of our constructed trajectory dataset with existing citation datasets, including RefSeer \cite{Huang2014RefSeerAC} and S2ORC \cite{Lo2020S2ORCTS}.}
    \label{tab:dataset_comparison}
\end{table}

\begin{table}[!t]
\centering
\small
\begin{tabular}{@{} >{\raggedright\arraybackslash}m{1.4cm} @{\hspace{4pt}} c @{\hspace{4pt}} m{5.1cm} @{}}
\toprule
\textbf{Category} & \textbf{Abbr.} & \textbf{Description} \\
\midrule
Background & BG & Provides background knowledge or information for the research domain. \\ \addlinespace
Motivation \& Gap & MG & Highlights research motivations or identifies gaps in existing literature. \\ \addlinespace
Comparison & CMP & Compares the current work or other baselines with specific prior methods. \\ \addlinespace
Basis \& Support & BS & Provides theoretical foundations or concepts supporting the current study. \\ \addlinespace
Method \& Data & MTD & Refers to specific methods, algorithms, tools, or datasets utilized in the research. \\ \addlinespace
Validation & VLD & Provides empirical evidence, results, or evaluation metrics to validate claims. \\ \addlinespace
Evolution & EV & Describes the historical evolution or development of a methodology. \\ \addlinespace
Critique & CRT & Expresses critiques, conflicts, or opposing views regarding prior work. \\
\bottomrule
\end{tabular}
\caption{Detailed descriptions of the 8-category Citation Intent Taxonomy (CAP-8).}
\label{tab:cap8_description}
\end{table}

\subsection{Citation Location Perception Data}
We formulate citation location perception as a sequence restoration task.
Given a paragraph with all original citations stripped, the task requires predicting a unified \texttt{[\#CITE\#]} marker at every legitimate citation position.
Beyond position recovery, we use DeepSeek-V3 to annotate the necessity of each ground-truth marker, classifying it as either \textit{Mandatory} (e.g., references to specific methods, algorithms, or datasets) or \textit{Optional} (e.g., general domain background).
By jointly supervising position and necessity, the dataset provides a more precise and context-aware location signal.
To support this task, we construct 31{,}118 samples (comprising 28{,}006 for training and 3{,}112 for testing) pairing the stripped input text with target outputs that mark every citation position together with its necessity label.
A complete example is shown in Appendix~\ref{sec:appendix_location}.

\begin{lstlisting}[captionpos=b, breakindent=0pt, caption={An example of the synthesized training data for QueryPlanner.}, label={lst:tool2_example}]
@@[Index 0]@@
!!<reasoning>!!
The purpose of this citation @@is to provide an example of@@ how testing can reveal biases or spurious correlations learned by a model, supporting the claim about the value of invariance testing. !!</reasoning>!!
!!<keywords>!!
[``spurious correlations'', ``text classification'', ``counterfactuals'', ``robustness''] !!</keywords>!!
\end{lstlisting}

\subsection{Query Planning Data}

Moving beyond citation location, the QueryPlanner dataset focuses on semantic intent.
Its objective is to bridge the vocabulary gap between a paragraph's vague local context and the precise metadata of the target reference.
Given a paragraph and a target citation position, the task is to generate both a reasoning chain that infers the citation intent and a keyword set that can retrieve the target reference.
We isolate 7{,}079 high-density paragraphs (comprising 6{,}779 for training and 300 for testing) whose BibTeX entries contain complete titles and abstracts, and use DeepSeek-V3 in a posterior setting: with full access to the target reference's metadata, it retroactively deduces a logical reasoning path (\texttt{<reasoning>}) and an optimal keyword set (\texttt{<keywords>}) for each citation (Listing~\ref{lst:tool2_example}).
This posterior view yields keywords that effectively retrieve the target.

To verify keyword quality, we run a cross-model consistency check on 100 random samples.
We substitute the keywords in 50 of these samples with keywords from unrelated instances to form negatives, and five LLMs binary-classify whether the keywords align with the target reference.
The 90.6\% average accuracy (Cohen's $\kappa = 0.805$) confirms the reliability of the synthesized data.
To analyze the logical distribution of training data, we run LLM-based motivation classification over 82{,}881 citation nodes; allowing secondary motivation labels expands the total to 93{,}403.
This yields the 8-category Citation Intent Taxonomy (CAP-8), with category definitions in Table~\ref{tab:cap8_description} and the corpus distribution in the intent panel of Figure~\ref{fig:data_source}.

\subsection{Reflective Trajectory Synthesis}
\begin{figure}[!t]
    \centering
    \includegraphics[width=\linewidth]{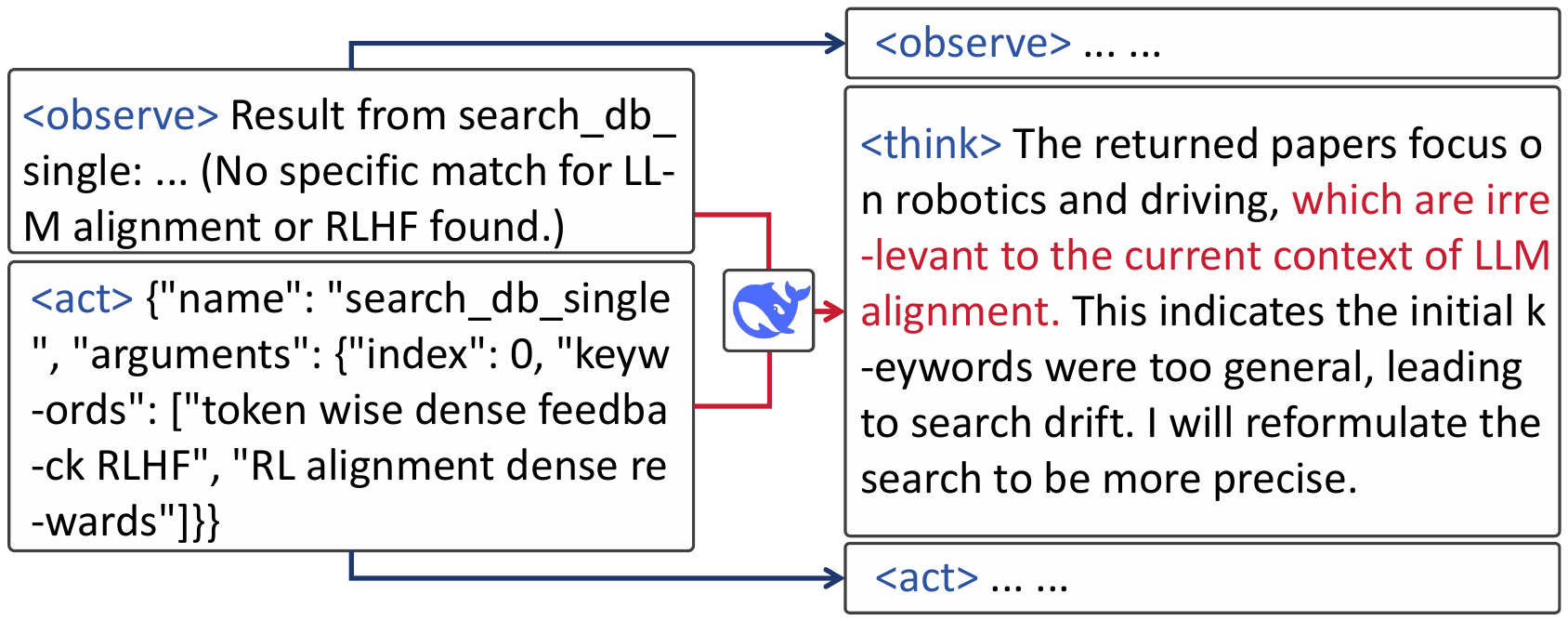}
    \caption{Data Generation Pipeline for Trajectory Thinking Process.
    }
    \label{fig:thinkdata}
\end{figure}

The third subset trains Master Brain to orchestrate the workflow and recover from retrieval failures.
Each sample is a multi-turn trajectory in an observe-think-act format: the agent observes the environment state (\texttt{<observe>}), articulates a plan (\texttt{<think>}), and issues an action (\texttt{<act>}) such as a tool call or a final citation choice.
To synthesize supervision, we fix \texttt{<observe>} and \texttt{<act>} from successful executions and use DeepSeek-V3 to retroactively deduce the connecting \texttt{<think>} (Figure~\ref{fig:thinkdata}).
We construct 2{,}000 trajectories in total: 500 ``gold'' paths and 1{,}500 reflection-enhanced variants.

For the gold paths, we inject real distractors from the local database so that the model must state within \texttt{<think>} why it prefers the target reference over each alternative.
For the reflection-enhanced variants, we inject vague initial keywords to simulate retrieval drift (e.g., returning irrelevant survey papers); the model must diagnose the failure, refine keywords from broad domains to specific entities, and issue a secondary retrieval before succeeding.
A complete example is in Appendix~\ref{sec:appendix_trajectory}.

To guarantee robust generalization and prevent data leakage, we construct a separate test set for our end-to-end citation prediction evaluation.
Specifically, we randomly sample 50 paragraphs from CVPR'25, EMNLP'24, ICCV'23, and NeurIPS'25, yielding 200 paragraphs in total.
This evaluation set is strictly disjoint from our training data pool and contains 1{,}023 unique target test papers.
An audit confirms that none of the 2{,}640 training citation decisions involved any of the 1{,}023 test papers.

\begin{figure*}[!t]
    \centering
    \includegraphics[width=1\textwidth]{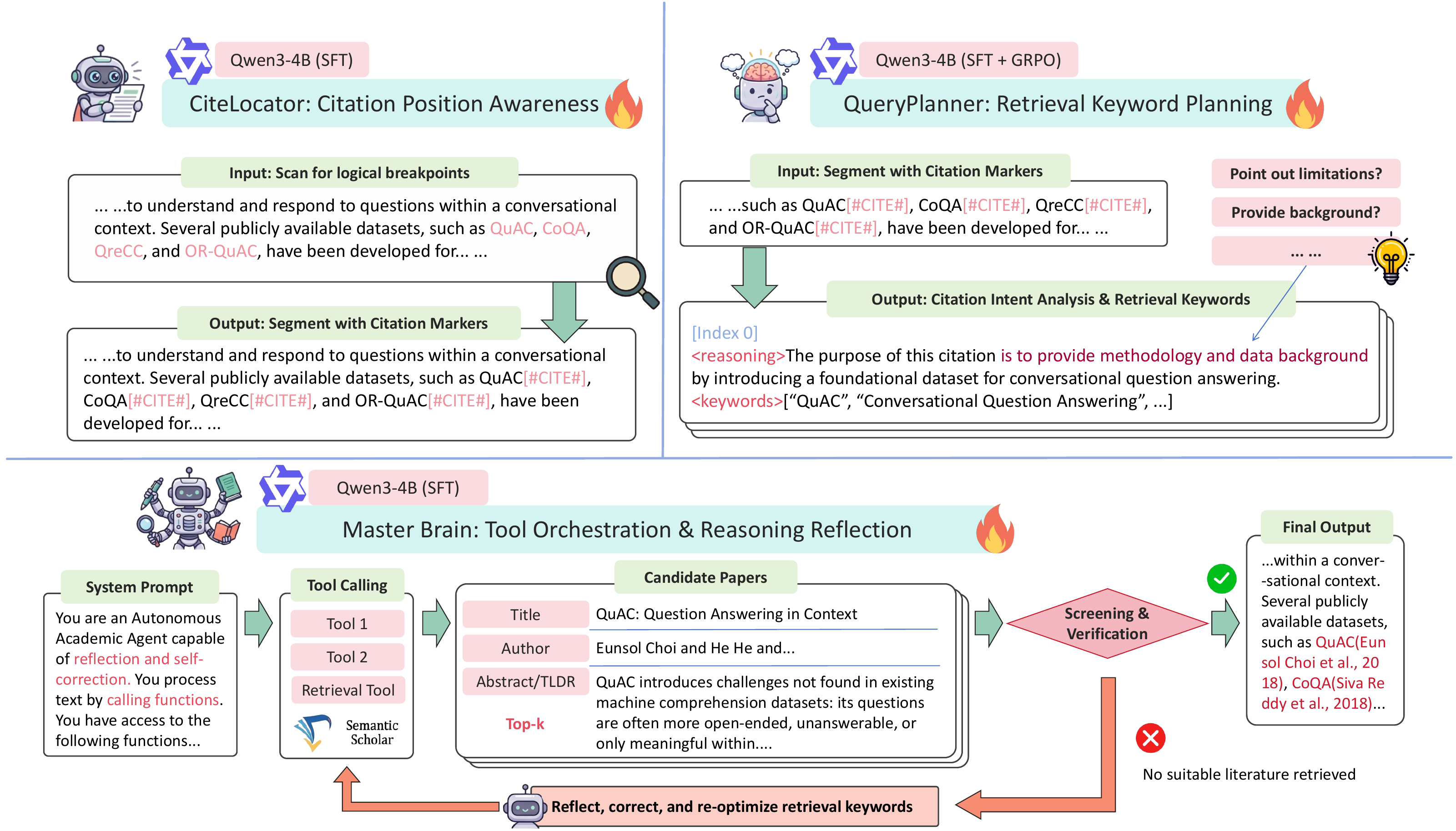}
    \caption{The overall architecture of our proposed \textbf{ReCite} framework. The system decouples the automatic citation process into three iterative stages: citation location perception (CiteLocator), intent-aware query planning (QueryPlanner), and task scheduling with reflective verification orchestrated by the Master Brain.
    }
    \label{fig:overall_architecture}
\end{figure*}

\section{Method}
\label{sec:method}

Figure~\ref{fig:overall_architecture} illustrates the framework of ReCite, and we detail each module below.

\subsection{Logical Breakpoint Perception}
CiteLocator predicts citation boundaries and their necessity labels (Mandatory or Optional).
We fine-tune Qwen3-4B~\cite{yang2025qwen3technicalreport} via SFT on the location perception data, registering the unified \texttt{[\#CITE\#]} marker as a new tokenizer entry so that it is not fragmented during encoding.
Because citation tokens are extremely sparse compared to regular text tokens, standard cross-entropy collapses to the majority class; we mitigate this with an asymmetric cross-entropy that places heavier loss weight on citation marker tokens, penalizing missed or hallucinated boundaries.

\subsection{Intent-Aware Query Planning}

QueryPlanner bridges the local context and precise retrieval metadata by generating a reasoning chain and a keyword set per citation position.
We first fine-tune Qwen3-4B via SFT on the query planning data, then apply GRPO~\cite{shao2024deepseekmathpushinglimitsmathematical} to elicit deeper reasoning.
For each query, GRPO samples a group of $G=4$ candidate outputs and updates the policy with their relative advantages, eliminating the need for a separate value network.
The composite reward combines three components: \textit{format compliance}, which penalizes deviations from the required \texttt{<reasoning>}/\texttt{<keywords>} layout; \textit{entity recall}, against the target reference; and \textit{intent alignment}, against the ground-truth intent.

\subsection{Task Orchestration}

Master Brain is the central orchestrator of the agentic workflow.
Like the other two modules, it is built on Qwen3-4B and trained via SFT on the reflective trajectory data to follow an observe-think-act state machine.
At inference, Master Brain first invokes CiteLocator to identify all citation positions in the draft.
For each position, it calls QueryPlanner to generate search keywords, queries the Semantic Scholar API to fetch candidate metadata (title, abstract, authors, year, citation count), and verifies whether the evidence supports the local claim.
When no candidate provides consistent evidence, the agent enters a reflective re-retrieval loop: it emits a failure-attribution \texttt{<think>}, refines keywords from broad domains to specific entities, and issues a secondary retrieval.
Because retrieval queries hit authentic databases, fabrication is impossible by construction; the residual challenge of misattribution is directly targeted by this verification and self-correction loop.

\section{Experiments}
\label{sec:experiments}

\subsection{Settings}

\begin{table*}[!t]
\centering
\small
\resizebox{\textwidth}{!}{
\begin{tabular}{llcccccccccc}
\toprule
\multirow{2}{*}{\textbf{Method}} & \multirow{2}{*}{\textbf{Size}} & \multicolumn{3}{c}{\textbf{Overall-Strictly (\%)}} & \multicolumn{3}{c}{\textbf{Lenient Evaluation (\%)}} & \multicolumn{3}{c}{\textbf{Position-Only (\%)}} \\
\cmidrule(lr){3-5} \cmidrule(lr){6-8} \cmidrule(lr){9-11}
& & \textbf{P} & \textbf{R} & \textbf{F1} & \textbf{P} & \textbf{R} & \textbf{F1} & \textbf{P} & \textbf{R} & \textbf{F1} \\
\midrule
\rowcolor[HTML]{F3F3F3} \multicolumn{11}{l}{\textit{Models via zero-shot prompt pipeline}} \\
GLM-4-Plus \cite{glm2024chatglmfamilylargelanguage} & --- & 9.83 & 10.27 & 10.05 & 17.22 & 17.99 & 17.60 & 47.08 & 49.19 & 48.11 \\
Mimo-V2.5-Pro \cite{mimo_xiaomi} & --- & 26.40 & 27.72 & 27.04 & 29.19 & 30.63 & 29.89 & 75.07 & 78.78 & 76.88 \\
GPT-4o-mini \cite{openai2024gpt4ocard} & --- & 7.20 & 8.65 & 7.86 & 12.73 & 15.29 & 13.89 & 44.89 & 53.90 & 48.98 \\
GPT-5.1-Chat \cite{singh2025openaigpt5card} & --- & 8.99 & 12.12 & 10.32 & 14.48 & 19.54 & 16.63 & 35.32 & 47.64 & 40.57 \\
Kimi-k2-preview \cite{kimiteam2026kimik2openagentic} & --- & 19.06 & 21.24 & 20.09 & 24.32 & 27.10 & 25.64 & 55.86 & 62.24 & 58.88 \\
Qwen3.6-Plus \cite{yang2025qwen3technicalreport} & --- & 28.09 & 35.44 & 31.34 & 31.09 & 39.23 & 34.69 & 68.60 & 86.50 & 76.52 \\
Qwen3.6-27B \cite{yang2025qwen3technicalreport} & 27B & 22.89 & 29.62 & 25.83 & 25.58 & 33.08 & 28.85 & 62.34 & 80.48 & 70.26 \\
DeepSeek-V4-Flash \cite{deepseekai2026deepseekv4} & 284B & 31.23 & 35.21 & 33.10 & 34.52 & 38.89 & 36.57 & 72.95 & 82.11 & 77.26 \\
DeepSeek-V4-Pro \cite{deepseekai2026deepseekv4} & 1.6T & 31.29 & 32.28 & 31.77 & 35.25 & 36.34 & 35.79 & 75.97 & 78.32 & 77.13 \\
\hdashline \noalign{\vskip 0.5ex}
\rowcolor[HTML]{F3F3F3} \multicolumn{11}{l}{\textit{General Agents (End to End via Mimo-V2.5-Pro)}} \\
Claude Code \cite{claude_code} & --- & 17.73 & 27.26 & 21.49 & 21.35 & 32.82 & 25.87 & 25.51 & 39.23 & 30.92 \\
Hermes Agent \cite{hermes_agent} & --- & 22.39 & 21.24 & 21.80 & 26.38 & 25.02 & 25.68 & 28.01 & 26.56 & 27.27 \\
OpenClaw \cite{openclaw} & --- & 18.09 & 7.03 & 10.12 & 25.25 & 9.81 & 14.13 & 29.42 & 11.43 & 16.46 \\
\hdashline \noalign{\vskip 0.5ex}
\rowcolor[HTML]{F3F3F3} \multicolumn{11}{l}{\textit{Ours}} \\
ReCite-Base & 4B & 36.92 & 35.75 & 36.33 & 43.46 & 42.08 & 42.76 & 89.80 & 86.11 & 87.53 \\
ReCite-SFT & 4B &  37.48 &  37.45 &  37.47 &  47.14 &  47.07 &  47.10 & 89.80 &  \textbf{89.66} & \textbf{89.73} \\
ReCite-SFT(CAP-8) & 4B & \textbf{39.22} & \textbf{39.07} & \textbf{39.15} & \textbf{55.27} & \textbf{55.02} & \textbf{55.14} &  \textbf{89.92} &  89.51 &  89.71 \\
\bottomrule
\end{tabular}
}
\caption{End-to-end citation prediction evaluation results. We report Precision (P), Recall (R), and F1-score across three metrics: Strict End-to-End, Lenient Evaluation, and Position-Only. CAP-8 denotes the model fine-tuned with citation taxonomy information.
Bold indicates the best result in each column.
}
\label{tab:evaluation_results}
\end{table*}

\begin{table*}[!t]
\centering
\small
\resizebox{\textwidth}{!}{
\begin{tabular}{llcccccccccc}
\toprule
\multirow{2}{*}{\textbf{Model}} & \multirow{2}{*}{\textbf{Size}} & \multirow{2}{*}{\textbf{Strategy}} & \multicolumn{3}{c}{\textbf{Mandatory (\%)}} & \multicolumn{3}{c}{\textbf{Optional (\%)}} & \multicolumn{3}{c}{\textbf{Overall (\%)}} \\
\cmidrule(lr){4-6} \cmidrule(lr){7-9} \cmidrule(lr){10-12}
& & & \textbf{P} & \textbf{R} & \textbf{F1} & \textbf{P} & \textbf{R} & \textbf{F1} & \textbf{P} & \textbf{R} & \textbf{F1} \\
\midrule
\rowcolor[HTML]{F3F3F3} \multicolumn{12}{l}{\textit{Baseline Models}} \\
GPT-5.1-Chat \cite{singh2025openaigpt5card} & --- & Direct & 22.01 & 59.72 & 32.17 & 9.21  & 58.50 & 15.92 & 27.73 & 59.39 & 37.81 \\
GPT-4o-mini \cite{openai2024gpt4ocard} & --- & Direct & 24.62 & 53.61 & 33.75 & 11.34 & 57.17 & 18.92 & 31.25 & 54.57 & 39.74 \\
Kimi-k2-preview \cite{kimiteam2026kimik2openagentic} & --- & Direct & 31.27 & 61.05 & 41.36 & 14.36 & 61.28 & 23.27 & 38.38 & 61.11 & 47.15 \\
MiniMax-M2.5 \cite{minimax_m25} & --- & Direct & 33.52 & 56.37 & 42.04 & 14.71 & 52.52 & 22.98 & 40.35 & 55.34 & 46.67 \\
GLM-4-flash \cite{glm2024chatglmfamilylargelanguage} & --- & Direct & 25.81 & 16.73 & 20.30 & 12.07 & 17.96 & 14.43 & 32.67 & 17.06 & 22.41 \\
Qwen3-Max \cite{yang2025qwen3technicalreport} & --- & Direct & 32.87 & 28.90 & 30.75 & 15.41 & 29.28 & 20.19 & 40.18 & 29.00 & 33.69 \\
Qwen3-Max \cite{yang2025qwen3technicalreport} & --- & CoT    & 34.43 & 28.42 & 31.14 & 16.00 & 28.09 & 20.39 & 41.71 & 28.33 & 33.74 \\
DeepSeek-V3 \cite{deepseekai2025deepseekv3technicalreport} & 671B & Direct & 30.91 & \textbf{62.66} & 41.40 & 13.91 & \textbf{61.65} & 22.70 & 37.85 & \textbf{62.39} & 47.12 \\
DeepSeek-V3 \cite{deepseekai2025deepseekv3technicalreport} & 671B & CoT    & 29.23 & 49.09 & 36.64 & 12.41 & 45.88 & 19.54 & 35.68 & 48.23 & 41.02 \\
Qwen3.5-27B \cite{yang2025qwen3technicalreport} & 27B  & Direct & 36.13 & 55.86 & 43.88 & 15.83 & 50.56 & 24.11 & 42.98 & 54.44 & 48.04 \\
Qwen3-4B \cite{yang2025qwen3technicalreport} & 4B   & Direct & 13.98 & 9.70  & 11.45 & 5.55  & 9.56  & 7.03  & 18.12 & 9.66  & 12.60 \\
Qwen3-4B \cite{yang2025qwen3technicalreport} & 4B   & CoT    & 8.98  & 3.99  & 5.52  & 2.59  & 2.92  & 2.74  & 11.12 & 3.70  & 5.55  \\
\midrule
\rowcolor[HTML]{F3F3F3} \multicolumn{12}{l}{\textit{Ours}} \\
\textbf{CiteLocator} & \textbf{4B} & \textbf{Direct} & \textbf{67.93} & 60.61 & \textbf{64.06} & \textbf{42.90} & 58.57 & \textbf{49.52} & \textbf{74.16} & 60.06 & \textbf{66.37} \\
\bottomrule
\end{tabular}
}
\caption{Citation location prediction results on the test set ($N=3,112$). We report Precision (P), Recall (R), and F1-score for Mandatory, Optional, and Overall categories. Bold indicates the best result in each column.
}
\label{tab:full_results}
\end{table*}

\paragraph{Baselines}
The base model for all three modules is Qwen3-4B-Instruct~\cite{yang2025qwen3technicalreport}.
We compare our framework against three categories of baselines: (i) zero-shot LLM pipelines that mirror our workflow;
(ii) \textbf{General Agents} (Claude Code~\cite{claude_code}, Hermes Agent~\cite{hermes_agent}, and OpenClaw~\cite{openclaw}) running end-to-end on a Mimo-V2.5-Pro inference engine; and (iii) heuristic and human baselines for location-only evaluation (GM-s2orc, GM-s2orc-H, and human experts~\cite{BUSCALDI2024103583}).
The full LLM baseline list is in Appendix~\ref{sec:appendix_setup}; detailed system prompts are in Appendices~\ref{sec:appendix_prompts}-\ref{sec:appendix_prompts_master}.

\paragraph{Evaluation Metrics}
Metrics strictly align with our architecture. CiteLocator uses Precision (P), Recall (R), and F1-score for \textit{Mandatory} and \textit{Optional} citations. QueryPlanner uses Token F1 (R-F1) for reasoning, alongside Token F1/Recall (K-F1, K-TR), Phrase Recall (K-PR), Jaccard Similarity (K-Jac), and an LLM-as-a-judge mechanism for keyword quality. End-to-end workflows are evaluated at three strictness levels: \textit{Overall-Strictly} (exact target match), \textit{Lenient} (relevant background/baselines), and \textit{Position-Only}.
Details on the evaluation metric definitions and calculations are provided in Appendix \ref{sec:appendix_metrics}.

\paragraph{Parameters}
For SFT, we use an 8-bit Paged AdamW optimizer (learning rate $2 \times 10^{-5}$, 3 epochs, cosine decay, 5\% warmup). Maximum sequence lengths are dynamically set: 1,024 for CiteLocator, 4,096 for QueryPlanner, and 9,216 for the Master Brain to preserve multi-turn trajectories. In the GRPO phase for QueryPlanner, the learning rate is $5 \times 10^{-7}$ with a sample group size of $G=4$.

\subsection{Main Results}

As shown in Table~\ref{tab:evaluation_results}, ReCite-SFT(CAP-8) achieves the highest Strict F1 of 39.15\%, outperforming every baseline including DeepSeek-V4-Pro and Qwen3.6-Plus, demonstrating the effectiveness of our decoupled agentic design.
This margin is especially notable despite the larger baselines likely benefiting from data leakage, since their pre-training corpora almost certainly cover our 2024-2025 test papers.
In addition, adding CAP-8 over ReCite-SFT yields a clear gain on both Strict and Lenient F1, showing that fine-grained intent supervision substantially improves candidate verification.
By comparison, General Agents (Claude Code, Hermes Agent, OpenClaw) fall well behind, indicating that black-box end-to-end agents cannot recover from retrieval drift, and that a task-specific agent design remains necessary in this setting.

While ReCite achieves state-of-the-art results, its Strict F1 (39.15\%) is heavily constrained by the severity of exact-match evaluation, which requires perfect joint accuracy across sequential stages and is confounded by subjective author biases.
Since scientific claims can often be validated by multiple appropriate papers, the 15.99\% gain in Lenient F1 (55.14\%) demonstrates ReCite's ability to discover logically supportive alternatives that are semantically valid but not explicitly cited by the original authors (a detailed qualitative case study is provided in Appendix~\ref{sec:appendix_case_study}).

\begin{table*}[!t]
\centering
\small
\resizebox{\textwidth}{!}{
\begin{tabular}{llccccccc}
\toprule
\multirow{2}{*}{\textbf{Model}} & \multirow{2}{*}{\textbf{Size}} & \multicolumn{2}{c}{\textbf{Reasoning}} & \multicolumn{5}{c}{\textbf{Keyword Extraction}} \\
\cmidrule(lr){3-4} \cmidrule(lr){5-9}
& & \textbf{R-F1 (\%)} & \textbf{R-Jdg} & \textbf{K-F1 (\%)} & \textbf{K-TR (\%)} & \textbf{K-PR (\%)} & \textbf{K-Jac (\%)} & \textbf{K-Jdg} \\
\midrule
\rowcolor[HTML]{F3F3F3} \multicolumn{9}{l}{\textit{Baseline Models}} \\
GPT-5.1-Chat \cite{singh2025openaigpt5card} & ---  & 52.25 & 7.48 & 39.05 & 44.03 & 12.16 & 7.87  & 6.26 \\
MiniMax-M2.5 \cite{minimax_m25} & ---  & 53.88 & 7.86 & 41.19 & 47.19 & 13.73 & 8.75  & 6.53 \\
Qwen3-Max \cite{yang2025qwen3technicalreport} & ---  & 53.91 & 7.33 & 37.01 & 36.56 & 15.38 & 10.83 & 5.95 \\
GLM-4-flash \cite{glm2024chatglmfamilylargelanguage} & ---  & 21.73 & 3.04 & 16.09 & 19.95 & 6.65  & 3.98  & 2.63 \\
Kimi-k2-preview \cite{kimiteam2026kimik2openagentic} & ---  & 54.12 & 7.75 & 45.80 & \textbf{55.30} & \textbf{25.45} & \textbf{16.42} & \textbf{7.12} \\
DeepSeek-V3 \cite{deepseekai2025deepseekv3technicalreport} & 671B & \textbf{58.63} & \textbf{7.88} & \textbf{46.25} & 49.18 & 22.59 & 15.13 & 6.65 \\
Qwen3.5-27B \cite{yang2025qwen3technicalreport} & 27B  & 50.55 & 7.56 & 40.99 & 43.62 & 16.73 & 11.14 & 6.33 \\
\hdashline \noalign{\vskip 0.5ex}
\rowcolor[HTML]{F3F3F3} \multicolumn{9}{l}{\textit{Ours}} \\
QueryPlanner(SFT) & 4B   & 56.12 & 7.21 & 37.74 & 43.56 & 16.87 & 10.38 & 5.93 \\
QueryPlanner(SFT+GRPO) & 4B & 57.81 & 7.64 & 39.06 & 42.47 & 17.85 & 11.00 & 6.44 \\
\bottomrule
\end{tabular}
}
\caption{Comprehensive evaluation results about intent-aware query planning on the test set ($N=300$). Bold indicates the best result in each column.}
\label{tab:full_citation_results}
\end{table*}

\begin{table}[!t]
\centering
\small
\begin{tabular}{llccc}
\toprule
 \textbf{Type} & \textbf{Method} & \textbf{P (\%)} & \textbf{R (\%)} & \textbf{F1 (\%)} \\
\midrule
\multirow{4}{*}{\makecell[l]{\textbf{Coarse-}\\\textbf{grained}}}
  & Scientist     & 76.2 & 88.4 & 81.8 \\
  & GM-s2orc      & 78.2 & 78.2 & 78.2 \\
  & GM-s2orc-H    & 80.2 & 82.6 & 81.4 \\
  & CiteLocator   & 66.7 & 87.0 & 75.5 \\
\midrule
\multirow{4}{*}{\makecell[l]{\textbf{Fine-}\\\textbf{grained}}}
  & Scientist     & 57.5 & 66.6 & 61.7 \\
  & GM-s2orc      & 44.9 & 44.9 & 44.9 \\
  & GM-s2orc-H    & 49.2 & 50.7 & 50.0 \\
  & CiteLocator   & 46.7 & 60.9 & 52.8 \\
\bottomrule
\end{tabular}
\caption{Performance comparison with existing baselines on the citation prediction task. The ``-H'' suffix indicates methods utilizing extra heuristic rules.}
\label{tab:comparison_with_baselines}
\end{table}

\subsection{Module-Level Evaluation}

Beyond end-to-end performance, we also evaluate CiteLocator and QueryPlanner individually against module-specific baselines.

\paragraph{CiteLocator.}
On location prediction (Table~\ref{tab:full_results}), CiteLocator outperforms all LLM baselines by a wide margin, reaching Overall F1 of 66.37\% versus 48.04\% for the next-best baseline (Qwen3.5-27B), with consistent gains across both Mandatory and Optional categories.
On the out-of-domain BUSCALDI benchmark (Table~\ref{tab:comparison_with_baselines}), which categorizes tasks into coarse-grained (sentence-level necessity) and fine-grained (token-level placement) predictions, CiteLocator significantly narrows the gap to human experts on exact location prediction. By outperforming GM-s2orc-H on Fine-grained, it demonstrates that our learned rhetorical signals generalize effectively beyond the training corpus.

\paragraph{QueryPlanner.}
On reasoning and keyword generation (Table~\ref{tab:full_citation_results}), QueryPlanner-SFT alone reaches Token F1 of 56.12\%, approaching the 58.63\% of DeepSeek-V3 (671B) and clearly beating the 27B open-source baseline.
Adding GRPO further lifts reasoning quality and most keyword metrics, with the composite reward (format + entity recall + intent alignment) pushing the model from surface copying to deeper entity grounding.

\subsection{Ablation Study}
To validate our decoupled framework, we individually replace CiteLocator, QueryPlanner, and the Master Brain with the Qwen3-4B-Instruct baseline (Table \ref{tab:ablation_study}). Replacing CiteLocator causes a severe drop in Strict F1 to 2.75\%, proving that logical breakpoint perception is the foundational prerequisite. Similarly, the absence of QueryPlanner lowers Strict F1 to 15.77\%, highlighting the necessity of intent-aware query planning. Also, removing the Master Brain's reflective verification degrades the Lenient F1 from 47.10\% to 42.76\%, indicating its crucial role in filtering out noise. Ultimately, the complete system achieves optimal performance, demonstrating that these modules are mutually reinforcing and indispensable.

\begin{table}[!t]
\centering
\scriptsize
\setlength{\tabcolsep}{5pt}
\begin{tabular}{l l@{~}l l@{~}l l@{~}l}
\toprule
\textbf{Method} & \multicolumn{2}{l}{\textbf{P (\%)}} & \multicolumn{2}{l}{\textbf{R (\%)}} & \multicolumn{2}{l}{\textbf{F1 (\%)}} \\
\midrule
\rowcolor{gray!15} \multicolumn{7}{l}{\textbf{Overall-Strictly}} \\
\midrule
ReCite & \textbf{37.48} & & \textbf{37.45} & & \textbf{37.47} & \\
w/o CiteLocator & 2.98 & \textcolor{red}{\tiny(-34.50)} & 2.55 & \textcolor{red}{\tiny(-34.90)} & 2.75 & \textcolor{red}{\tiny(-34.72)} \\
w/o QueryPlanner & 15.79 & \textcolor{red}{\tiny(-21.69)} & 15.75 & \textcolor{red}{\tiny(-21.70)} & 15.77 & \textcolor{red}{\tiny(-21.70)} \\
w/o Master Brain & 36.92 & \textcolor{red}{\tiny(-0.56)} & 35.75 & \textcolor{red}{\tiny(-1.70)} & 36.33 & \textcolor{red}{\tiny(-1.14)} \\
\midrule
\rowcolor{gray!15} \multicolumn{7}{l}{\textbf{Lenient Evaluation}} \\
\midrule
ReCite & \textbf{47.14} & & \textbf{47.07} & & \textbf{47.10} & \\
w/o CiteLocator & 7.05 & \textcolor{red}{\tiny(-40.09)} & 6.02 & \textcolor{red}{\tiny(-41.05)} & 6.49 & \textcolor{red}{\tiny(-40.61)} \\
w/o QueryPlanner & 30.88 & \textcolor{red}{\tiny(-16.26)} & 30.81 & \textcolor{red}{\tiny(-16.26)} & 30.85 & \textcolor{red}{\tiny(-16.25)} \\
w/o Master Brain & 43.46 & \textcolor{red}{\tiny(-3.68)} & 42.08 & \textcolor{red}{\tiny(-4.99)} & 42.76 & \textcolor{red}{\tiny(-4.34)} \\
\bottomrule
\end{tabular}
\caption{Ablation study on different components.}
\label{tab:ablation_study}
\end{table}

\begin{figure*}[!t]
    \centering
    \includegraphics[width=1\textwidth]{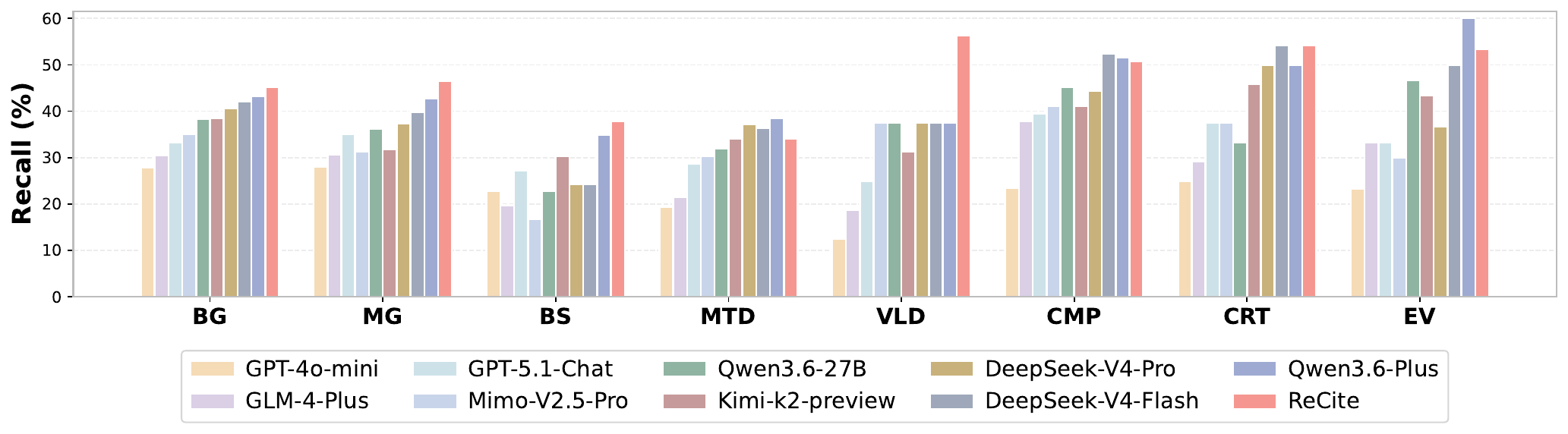}
    \caption{Performance comparison of various models across different citation intent categories. The evaluation is based on the Recall metric within the CAP-8 taxonomy.
    }
    \label{fig:intent_performance_comparison}
\end{figure*}

\begin{figure}[!t]
    \centering
    \includegraphics[width=1\linewidth]{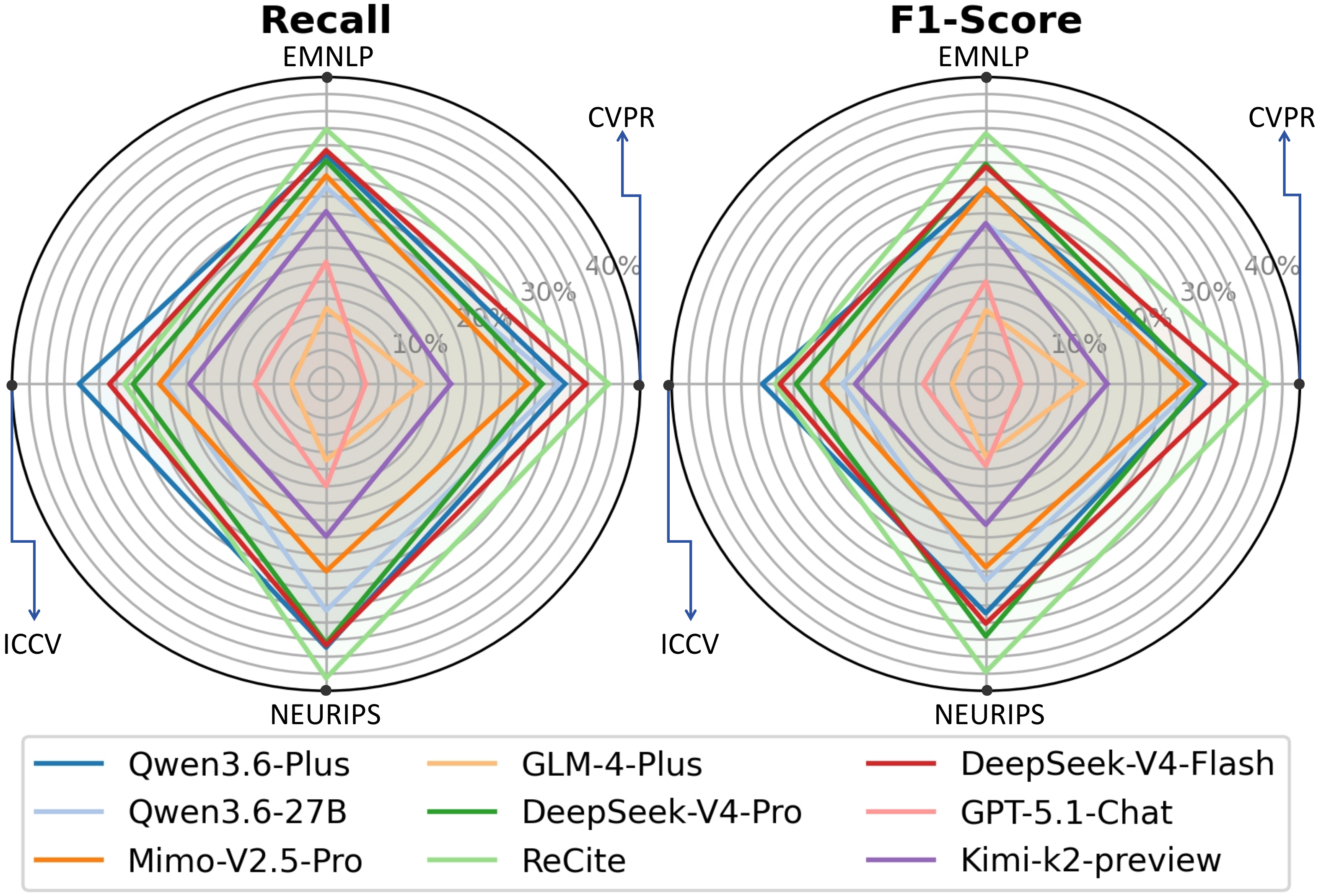}
    \caption{
    Performance comparison of different models based on the Recall and F1-Score.
    }
    \label{fig:f1_radar_chart}
\end{figure}

\subsection{In-depth Analysis}
\paragraph{Analysis of Cross-Venue Data Robustness}
Figure~\ref{fig:f1_radar_chart} breaks down end-to-end F1 by the source conference of each test paragraph.
ReCite stays balanced across all four sub-domains, while several general-purpose baselines fluctuate or skew toward specific venues.
This suggests that ReCite captures domain-general rhetorical signals rather than surface vocabulary overlap, supporting cross-domain transfer within computer science.

\paragraph{Analysis of Intent-Level Recall}
Figure~\ref{fig:intent_performance_comparison} reports recall across the CAP-8 intent categories.
ReCite achieves higher recall than most baselines across categories, indicating that explicit intent supervision helps the agent align retrieval with author motivation rather than relying on semantic overlap.

\begin{figure}[!t]
    \centering
    \includegraphics[width=1\linewidth]{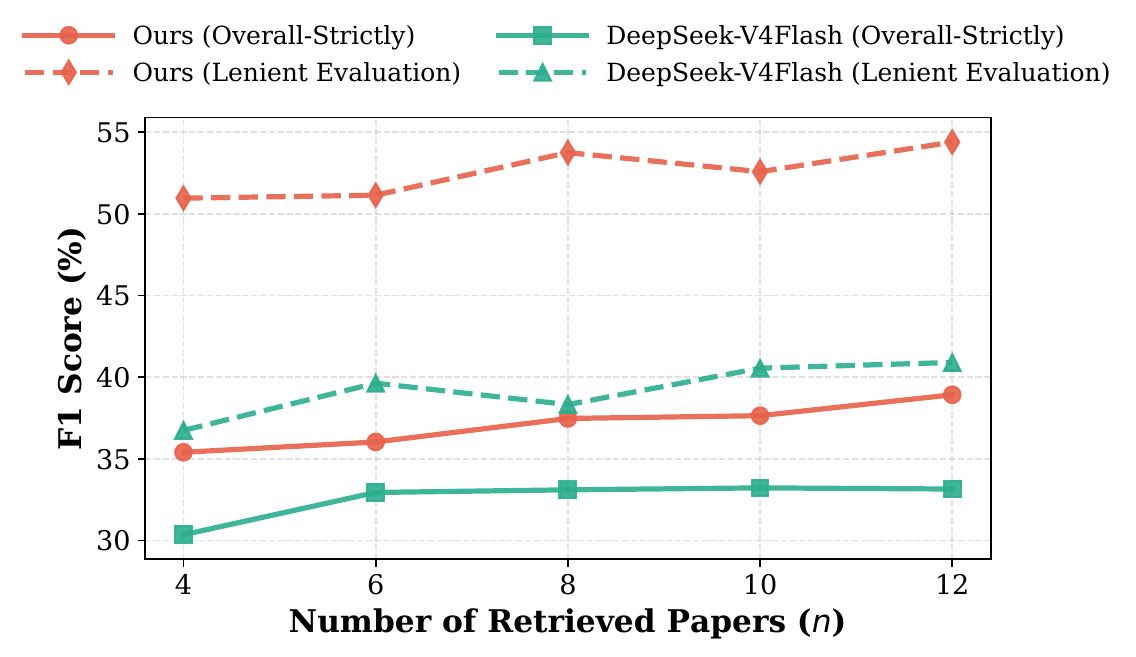}
    \caption{Performance variation of the proposed system and the baseline across different numbers of retrieved candidate papers (Top-k).
    }
    \label{fig:limit}
\end{figure}

\paragraph{Analysis of Retrieval Candidate Quantity}

As k varies from 4 to 12 (Figure~\ref{fig:limit}), the end-to-end F1 scores of both ReCite and the DeepSeek-V4-Flash baseline improve up to k = 8, after which further improvements are limited by context noise.
Moreover, a larger $k$ inflates input length and inference cost; we therefore fix $k=8$ as the optimal trade-off between evidence sufficiency and latency.

\section{Conclusion}
\label{sec:conclusion}
In this paper, we introduced ReCite, a decoupled agentic reasoning framework for faithful citation. ReCite integrates citation location perception, intent-aware query planning, and reflective verification to address misattribution while eliminating fabricated references through retrieval from authentic academic databases. To train these capabilities, we constructed a large-scale reasoning-oriented dataset from \LaTeX{} sources and synthesized supervision for location prediction, citation intent inference, keyword planning, and self-corrective trajectories. Extensive experiments show that ReCite achieves stronger strict citation accuracy than substantially larger generative models and general-purpose agents, demonstrating the effectiveness of our lightweight modules. Ablation and intent-level analyses confirm that each component contributes to robust citation grounding. By establishing citation as verifiable reasoning, ReCite provides a practical foundation for AI-assisted academic writing that is more faithful, accountable, and trustworthy.

\section*{Limitations}
While \textbf{ReCite} demonstrates strong performance in specific citation tasks, several limitations remain. First, our specialized lightweight model inherently trails massive general-purpose LLMs in open-ended reasoning outside the targeted workflow. Second, the system currently operates exclusively on uni-modal text. Since real-world citations often rely on charts or pseudocode, future work will integrate Vision-Language Models (VLMs) to support multimodal verification.
Third, because the framework was trained entirely on computer science papers, its cross-disciplinary generalization (e.g., to the humanities or biomedical fields) remains unexplored and requires systematic evaluation.

\section*{Ethical Considerations}
Our research directly addresses a critical ethical issue in AI-assisted scientific writing: the fabrication of references, which severely misleads readers and undermines academic integrity. By enforcing a rigorous verification loop grounded in real-world academic databases, our framework mitigates the propagation of hallucinated knowledge.

We recognize the ethical concerns surrounding data privacy and intellectual property. Academic research often involves highly confidential, unpublished data, and relying on closed-source, cloud-based LLMs poses inherent compliance and data leakage risks. A core motivation for developing our framework on a lightweight 4B parameter model is to facilitate future model quantization and pruning. This paves the way for fully offline, on-device deployment, ensuring that researchers can leverage agentic citation assistance without compromising sensitive intellectual property.

\section*{Acknowledgements}
We acknowledge the use of AI assistants during the writing and coding phases of this research. These tools were utilized solely to refine linguistic phrasing and assist with coding, while all scientific concepts, experimental designs, and final analyses were developed entirely by the authors.

\bibliography{custom}

\clearpage
\newpage

\appendix
\section{Experiment Settings}
\label{sec:appendix_setup}

\paragraph{Zero-shot LLM Baselines.}
For end-to-end and module-level evaluation we benchmark against proprietary models (GPT-4o-mini~\cite{openai2024gpt4ocard}, GPT-5.1-Chat~\cite{singh2025openaigpt5card}, Kimi-k2-preview~\cite{kimiteam2026kimik2openagentic}, GLM-4-Plus, GLM-4-Flash~\cite{glm2024chatglmfamilylargelanguage}, Qwen3.6-Plus, Qwen3-Max~\cite{yang2025qwen3technicalreport}, Mimo-V2.5-Pro~\cite{mimo_xiaomi}, MiniMax-M2.5~\cite{minimax_m25}) and open-source LLMs (Qwen3.5-27B, Qwen3-4B~\cite{yang2025qwen3technicalreport}, DeepSeek-V3~\cite{deepseekai2025deepseekv3technicalreport}, DeepSeek-V4-Flash, DeepSeek-V4-Pro~\cite{deepseekai2026deepseekv4}), all evaluated zero-shot under a prompt pipeline mirroring our workflow.

\paragraph{General Agents.}
The three end-to-end agents (Claude Code, Hermes Agent, OpenClaw) receive a single end-to-end citation objective and orchestrate their own tool calls; for fair comparison, all three run on Mimo-V2.5-Pro.

\paragraph{Heuristic and Human Baselines.}
For location-only evaluation, we compare against GM-s2orc, a GPT-2 based generative model fine-tuned on the s2orc dataset, and its variant GM-s2orc-H, which incorporates additional post-hoc NLP heuristics. Scientist denotes human annotator performance from~\cite{BUSCALDI2024103583}. To evaluate out-of-domain generalization, we utilize their introduced gold standard dataset, which comprises 133 sentences randomly sampled from recent arXiv papers and strictly annotated by three senior researchers.

\paragraph{Implementation Details.}
We train and evaluate ReCite on NVIDIA RTX 4090 D and A800 GPUs, with DeepSpeed for distributed training and vLLM (PagedAttention) for high-throughput inference.

\section{Details of Evaluation Metrics}
\label{sec:appendix_metrics}
We detail the evaluation metrics, strictly aligning with the three modules of the ReCite framework.

\subsection{Main Results}
The end-to-end evaluation assesses the final citation mapping at three strictness levels:
\begin{itemize}
    \item \textbf{Overall-Strictly:} Requires both correct identification of the citation position AND successful retrieval of the exact ground-truth paper.
    \item \textbf{Lenient Evaluation:} Accepts highly relevant substitute papers, acknowledging that multiple papers can logically support a claim. This is assessed via an \textbf{LLM-as-a-Judge} that outputs a binary score (1 for academically appropriate, 0 otherwise) based on local context and candidate metadata.
    \item \textbf{Position-Only:} Evaluates structural perception alone. A prediction is correct if it successfully identifies the logical breakpoint, regardless of the retrieved candidate.
\end{itemize}

\subsection{Citation Location Perception (CiteLocator)}
We evaluate the generated \texttt{[\#CITE\#]} placeholders against ground-truth breakpoints using standard Precision ($P$), Recall ($R$), and F1-score ($F1$):
\begin{equation}
    P = \frac{TP}{TP + FP}, \quad R = \frac{TP}{TP + FN}
\end{equation}
\begin{equation}
    F1 = \frac{2 \times P \times R}{P + R}
\end{equation}
These metrics are computed separately for \textit{Mandatory} (specific evidentiary backing) and \textit{Optional} (general background) citations.

\subsection{Intent-Aware Query Planning (QueryPlanner)}
To evaluate the QueryPlanner, we utilize the following metrics:
\begin{itemize}
    \item \textbf{Token Metrics (R-F1, K-F1, K-TR):} Unigram F1 overlap for the generated \texttt{<reasoning>} (R-F1) and \texttt{<keywords>} (K-F1), along with the keyword Token Recall (K-TR), evaluated against the ground-truth.
    \item \textbf{Phrase Recall (K-PR):} The proportion of ground-truth core entities successfully extracted in the \texttt{<keywords>} list.
    \item \textbf{Jaccard Similarity (K-Jac):} The intersection over union of the predicted ($K_{pred}$) and ground-truth ($K_{gt}$) keyword sets:
    \begin{equation}
        \text{K-Jac} = \frac{|K_{pred} \cap K_{gt}|}{|K_{pred} \cup K_{gt}|}
    \end{equation}
    \item \textbf{LLM-as-a-Judge:} An advanced LLM scores the keyword quality (0-10) based on logical soundness and retrieval feasibility.
\end{itemize}

\section{Data Examples and System Prompts}
\label{sec:appendix_prompts_examples}

\subsection{Examples of Location Perception Data}
\label{sec:appendix_location}

\begin{lstlisting}[breakindent=0pt, caption={Example input and output for citation location perception data.}]
!!input:!! Most state-of-the-art FSRL methods adopt a meta-learning framework, where a base model is pre-trained... they predominantly rely on KG embedding models like TransE or DistMult to exploit relational information.

!!output:!! Most state-of-the-art FSRL methods adopt a meta-learning framework @@[#CITE#]@@, where a base model is pre-trained... they predominantly rely on KG embedding models like TransE @@[#CITE#]@@ or DistMult @@[#CITE#]@@ to exploit relational information.
\end{lstlisting}

\subsection{Example of Reflective Trajectory Data}
\label{sec:appendix_trajectory}

\begin{lstlisting}[breakindent=0pt, caption={Example of multi-turn reflective trajectory data synthesis, demonstrating self-correction.}]
=== Turn 3 Assistant ===
!!<think>!!
The initial intent analysis suggested keywords such as reinforcement learning feedback''. I will use these to search for candidate papers for index 0.
!!</think>!!
!!<act>!!
{name'': search_db_single'',arguments'': {index'': 0,keywords'': [``reinforcement learning feedback'']}}
!!</act>!!

=== Turn 4 User ===
!!<observe>!!
Result from search_db_single: ... (No specific match for LLM alignment or RLHF found.)
!!</observe>!!

=== Turn 5 Assistant ===
!!<think>!!
The returned papers focus on robotics and driving, @@which are irrelevant to the current context@@ of LLM alignment. This indicates the initial keywords were too general, leading to search drift. I will reformulate the search to be more precise.
!!</think>!!
!!<act>!!
{name'':search_db_single'', arguments'': {index'': 0, keywords'': [token-wise dense feedback RLHF'', ``RL alignment dense rewards'']}}
!!</act>!!
\end{lstlisting}

\subsection{System Prompts for Pipeline Baselines}
\label{sec:appendix_prompts}

To ensure reproducibility and fair comparison, we provide the exact zero-shot system prompts used for our three-stage pipeline baselines.

\begin{lstlisting}[breakindent=0pt, caption={Zero-shot Prompt for Citation Location Perception (CiteLocator Pipeline).}]
!![System Prompt]!!
You are a professional academic copy-editor.

!![User Prompt]!!
Task: Act as a professional academic copy-editor. Insert the specified marker @@[#CITE#]@@ at places that require citations, such as when mentioning prior works, existing methods, algorithms, models, datasets or other places where citations are needed. 
Rules: Do not alter any original content. Do not add extra preamble. Output only the processed text.

!!Input Text:!!{text}
\end{lstlisting}

\begin{lstlisting}[breakindent=0pt, caption={Zero-shot Prompt for Intent-Aware Query Planning (QueryPlanner Pipeline).}]
@@You are an expert academic citation analyst.@@

!!### Task!!
Analyze the context of the given text. For each citation marker [#CITE#], infer what keywords should be used to search for the paper that needs to be cited. Each [#CITE#] marker in the input must correspond to one [Index n] in the output in order.

!!### Output Components!!
@@1. <reasoning>:@@ A 1-2 sentence logical analysis of the context explaining the core role of the citation. MUST start with the fixed phrase: ``The purpose of this citation is to...''
@@2. <keywords>:@@ 3-5 precise keywords to retrieve the target document.

!!### Constraints!!
- Sequential Indexing: Start with [Index 0] and increment by 1 for each marker.
- Output ONLY the structured content. Strictly NO conversational filler.

!!### Output Format!!
[Index X]
<reasoning>...</reasoning>
<keywords>[``keyword 1'', ``keyword 2'', ...]</keywords>
\end{lstlisting}

\begin{lstlisting}[breakindent=0pt, caption={Zero-shot Prompt for Candidate Verification (Search and Submit Pipeline).}]
!![System Prompt]!!
You are an academic paper selector. Your goal is to match the exact foundational paper needed for the specific citation context.

!![User Prompt]!!
You need to select the best citation for the marker [Index {index}] (which corresponds to the {index+1}-th [#CITE#] marker in the text).

!!ORIGINAL CONTEXT:!!
{marked_text}

!!Your reasoning for this citation was:!! {reasoning}

!!Here are the candidate papers from the database:!!
{candidate papers}

Evaluate carefully based on the original context. If you find the correct foundational paper, reply with ONLY its ID. If NONE of these papers match the context, reply with exactly ``NONE''.
\end{lstlisting}

\subsection{System Prompts for General Agents}
\label{sec:appendix_prompts_general}

\begin{lstlisting}[breakindent=0pt, caption={System Prompt for Autonomous General Agents (End-to-End via Mimo-V2.5-Pro).}]
You are an @@automatic citation assistant@@. Your capability is to identify places in given text paragraphs that require academic citations, autonomously construct keywords to retrieve real academic papers, and strictly follow the rules to add standard citations. It is strictly forbidden to search the original source article to copy existing real citations, or to generate fake papers.

!!STRICT CITATION RULES:!!
Citation Judgment: Identify positions that require citations, such as when mentioning prior works, existing methods, algorithms, models, datasets, or other places where citations are needed.
Literature Retrieval: Actively use keywords to retrieve real literature for matching citations.
Citation Placement Rule: Insert exactly one ordered citation mark (e.g., [1], [2], [3]...) for each independent sentence or academic claim. Multiple citations stacking at the same position are strictly prohibited.
Bibliography Format: Append an exact line titled ``Bibliography:'' at the very end of the modified full text. Each bibliography entry follows the fixed format: [Number] First Author, ``Exact Full Paper Title''
No Extra Content: Do not add any redundant explanations, comments, or modification notes in the output text.

!!FIXED OUTPUT FORMAT (MANDATORY):!!
Output a single-line valid JSON object for each processed entry, containing only two fields with no extra content:
{``text'':``Full modified text with inserted citations and final Bibliography section'', ``bibliography'':[``First Author, \``Exact Full Paper Title\'']}

!!HARD CONSTRAINTS:!!
Never modify the original input file in any way.
All cited literature must be real and retrievable; fake literature and copying original citations are strictly prohibited.
Strictly follow the ``one sentence, one citation'' rule; no citation stacking or redundant citation marking.
\end{lstlisting}

\subsection{System Prompt for Master Brain Agent}
\label{sec:appendix_prompts_master}

\begin{lstlisting}[breakindent=0pt, caption={System Prompt for the Master Brain Agent (Task Orchestration and Reflective Verification).}]
You are an Autonomous Academic Agent capable of reflection and self-correction. You process text by calling functions. You have access to the following functions: {functions}

!!CRITICAL RULES:!!
1. Sequential Workflow: mark_citations -> analyze_intents -> (Loop starts) -> search_db_single (Index 0) -> evaluate_and_submit_single (Index 0) -> search_db_single (Index 1) ... -> finalize_chunk.
2. You MUST process each index one by one. Do NOT search for Index 1 until you have submitted Index 0.
3. Once all indices reported by analyze_intents are submitted, call finalize_chunk.

!!OUTPUT FORMAT: !!
Every response MUST consist of two parts:
1. A reasoning process wrapped in <think>...</think> tags.
2. A function call wrapped in <act>...</act> tags.
\end{lstlisting}

\section{Case Study on the Multi-Option Reality in Citations}
\label{sec:appendix_case_study}

To illustrate the gap between Strict and Lenient metrics, we analyze a representative case from our test set.
Consider the context: \textit{``Compared with the global post-processing optimization used in previous work \texttt{[\#CITE\#]}, our method learns the descent direction...''}
While the original authors cited GLAMR (Yuan et al., 2022), ReCite retrieved HuMoR (Rempe et al., 2021), a prominent framework that also utilizes a gradient-based global optimization loop for 3D human motion estimation.
Both papers are academically appropriate and provide equivalent logical validation for the claim.

Under the Strict exact-match metric, retrieving HuMoR is penalized with a score of zero due to the string mismatch with the authors' specific bibliographic key.
In contrast, our Lenient Evaluation correctly identifies HuMoR as a valid semantic alternative.
This case highlights how Strict F1 significantly underrepresents ReCite's true citation faithfulness by ignoring acceptable multi-option academic realities and subjective author preferences.

\end{document}